%% file: CIKM_Preprint.tex
\documentclass[sigconf, nonacm=true]{acmart}

\AtBeginDocument{%
  \providecommand\BibTeX{{%
    \normalfont B\kern-0.5em{\scshape i\kern-0.25em b}\kern-0.8em\TeX}}}

\usepackage[noend]{algorithmic}
\usepackage{algorithm}
\usepackage{makecell}
\usepackage{balance} 
\input{packages}

\input{math_commands}
\long\def\comment#1{}

\newcommand\ALGPHASE[1]{%
\vspace*{-0.5\baselineskip}\hspace*{\dimexpr-2pt\relax}\rule{\linewidth}{0.4pt}%
\\\hspace*{-\algorithmicindent}\textbf{#1}%
\vspace*{-1.8\baselineskip}\\\hspace*{\dimexpr-2pt\relax}\rule{\linewidth}{0.4pt}%
}

\begin{document}

\title[Federated Deep Equilibrium Learning]{Federated Deep Equilibrium Learning: Harnessing Compact Global Representations to Enhance Personalization}

\author{Long Tan Le}
\email{long.le@sydney.edu.au}
\orcid{0000-0003-3284-1990}
\affiliation{%
  \institution{The University of Sydney}
  \city{Sydney}
  \state{NSW}
  \country{Australia}
  \postcode{2006}
}
\author{Tuan Dung Nguyen}
\email{joshtn@seas.upenn.edu}
\orcid{0000-0002-1105-005X}
\affiliation{%
  \institution{The University of Pennsylvania}
  \city{Philadelphia}
  \state{PA}
  \country{USA}
  \postcode{19104}
}

\author{Tung-Anh Nguyen}
\email{tung6100@uni.sydney.edu.au}
\orcid{0009-0000-0664-0218}
\affiliation{%
  \institution{The University of Sydney}
  \city{Sydney}
  \state{NSW}
  \country{Australia}
  \postcode{2006}
}

\author{Choong Seon	Hong}
\email{cshong@khu.ac.kr}
\orcid{0000-0003-3484-7333}
\affiliation{%
  \institution{Kyung Hee University}
  \city{Yongin-si}
  \country{Republic of Korea}
  \postcode{17104}
}

\author{Suranga	Seneviratne}
\email{suranga.seneviratne@sydney.edu.au}
\orcid{0000-0002-5485-5595}
\affiliation{%
  \institution{The University of Sydney}
  \city{Sydney}
  \state{NSW}
  \country{Australia}
  \postcode{2006}
}

\author{Wei Bao}
\email{wei.bao@sydney.edu.au}
\orcid{0000-0003-1874-1766}
\affiliation{%
  \institution{The University of Sydney}
  \city{Sydney}
  \state{NSW}
  \country{Australia}
  \postcode{2006}
}

\author{Nguyen H. Tran}
\email{nguyen.tran@sydney.edu.au}
\orcid{0000-0001-7323-9213}
\affiliation{%
  \institution{The University of Sydney}
  \city{Sydney}
  \state{NSW}
  \country{Australia}
  \postcode{2006}
}



\renewcommand{\shortauthors}{Long Tan Le et al.}

\input{./CIKMSections/abstract}







\maketitle

\input{./CIKMSections/intro}
\input{./CIKMSections/related_works}
\input{./CIKMSections/problem_formulation}
\input{./CIKMSections/experiments}

\input{./CIKMSections/conclusion}

\bibliographystyle{ACM-Reference-Format}
\balance
\bibliography{cikm_references}

\end{document}

%% file: packages.tex
\usepackage{microtype}

\usepackage{mathtools}
\usepackage{amsthm}
\usepackage{enumitem}

\usepackage[capitalize,noabbrev]{cleveref}

\makeatletter
\newcommand*{\addFileDependency}[1]{
  \typeout{(#1)}
  \@addtofilelist{#1}
  \IfFileExists{#1}{}{\typeout{No file #1.}}
}
\makeatother

\theoremstyle{plain}
\newtheorem{theorem}{Theorem}[section]

\newtheorem{lemma}[theorem]{Lemma}

\theoremstyle{definition}

\newtheorem{assumption}[theorem]{Assumption}
\theoremstyle{remark}
\newtheorem{remark}[theorem]{Remark}

\usepackage{hyperref}

\usepackage{url}
\usepackage{soul}
\usepackage{url}
\usepackage[utf8]{inputenc}

\usepackage{wrapfig}

\usepackage{setspace}
\input{math_commands.tex}
\crefformat{equation}{(#2#1#3)}
\crefrangeformat{equation}{(#3#1#4) to~(#5#2#6)}
\crefmultiformat{equation}{(#2#1#3)}%
{ and~(#2#1#3)}{, (#2#1#3)}{ and~(#2#1#3)}
\crefname{assumption}{assumption}{assumptions}
\crefname{Assumption}{Assumption}{Assumptions}

\usepackage{textcomp}
\usepackage{xcolor}

\usepackage{xspace}
\usepackage{subcaption}
\usepackage{bm}
\usepackage{multirow}

\makeatletter
\newcommand{\multilines}[1]{%
	\begin{tabularx}{\dimexpr\linewidth-\ALG@thistlm}[t]{@{}X@{}}
		#1
	\end{tabularx}
}
\makeatother

\newcommand{\OurAlg}{\ensuremath{\textsf{FeDEQ}}\xspace}
\Crefname{figure}{Fig.}{Figs.}
\Crefname{table}{Table}{Tables}
\Crefname{section}{Sec.}{Secs.}

\DeclarePairedDelimiterX{\norm}[1]{\lVert}{\rVert}{#1}
\DeclarePairedDelimiterX{\abs}[1]{\lvert}{\rvert}{#1}
\newcommand{\innProd}[1]{\left\langle#1\right\rangle}

\DeclareSymbolFont{extraup}{U}{zavm}{m}{n}
\DeclareMathSymbol{\varheart}{\mathalpha}{extraup}{86}

\newcommand{\defeq}{\vcentcolon=}

\DeclareUnicodeCharacter{FB01}{fi}



%% file: math_commands.tex
\def\eqref#1{(\ref{#1})}

\def\1{\bm{1}}

\def\RR{\mathbb{R}}

\def\gL{{\mathcal{L}}}

\def\gX{{\mathcal{X}}}
\def\gY{{\mathcal{Y}}}

\def\argmin{\text{arg\,min}}

%% file: CIKMSections/abstract.tex
\begin{abstract}
  Federated Learning (FL) has emerged as a groundbreaking distributed learning paradigm enabling clients to train a global model collaboratively without exchanging data. Despite enhancing privacy and efficiency in information retrieval and knowledge management contexts, training and deploying FL models confront significant challenges such as communication bottlenecks, data heterogeneity, and memory limitations. To comprehensively address these challenges, we introduce \OurAlg, a novel FL framework that incorporates deep equilibrium learning and consensus optimization to harness compact global data representations for efficient personalization. Specifically, we design a unique model structure featuring an equilibrium layer for global representation extraction, followed by explicit layers tailored for local personalization. We then propose a novel FL algorithm rooted in the alternating directions method of multipliers (ADMM), which enables the joint optimization of a shared equilibrium layer and individual personalized layers across distributed datasets. Our theoretical analysis confirms that \OurAlg converges to a stationary point, achieving both compact global representations and optimal personalized parameters for each client. Extensive experiments on various benchmarks demonstrate that \OurAlg matches the performance of state-of-the-art personalized FL methods, while significantly reducing communication size by up to 4 times and memory footprint by 1.5 times during training.
\end{abstract}

%% file: CIKMSections/intro.tex
\section{Introduction}
\label{sec:intro}

As the surge of Internet-connected devices leads to massive data generation, the shortcomings of centralized computing are becoming increasingly evident. Federated learning (FL) has emerged as a compelling countermeasure, facilitating data processing closer to its source. 
Unlike traditional machine learning methods that require data transfer to a central server, FL enables multiple clients to collaborate on training a unified model without exchanging data. Here, clients refine the model using their local data and transmit only the model's updates to a global coordinator for aggregation~\cite{fedavg}. 
Despite offering robust data privacy and responsiveness, the full realization of FL's potential is contingent upon overcoming three major challenges: (1) \textit{communication bottlenecks}, resulting from the need to transmit large model parameters between parties; (2) \textit{data heterogeneity} arises from the variety and diversity of the data held by clients, causing \textit{client drift} effects where individual model performances diverge; and (3) \textit{memory limitations} of clients' devices constrain the complexity and size of models that can be employed. 

Recent literature has presented diverse approaches to address the foregoing challenges. Communication and memory challenges have been addressed using techniques such as model compression~\citep{com1} to reduce the size of the model for transmission and adaptive gradients~\citep{comadaptive} for efficient parameter updates. Meanwhile, personalization has been seen as a key strategy to address data heterogeneity and client drift. Full model personalization methods, such as meta-learning~\citep{perfedavg, pfedme} and multi-task learning \citep{smith_federated_2018, Marfoq2021, ditto}, have shown efficiency in capturing unique local patterns. However, these techniques can be resource-intensive and may necessitate considerable communication overhead. Conversely, partial personalization approaches~\citep{fedrep, fedalt} offer a balanced solution by employing a shared representation complemented by personalized local models, enabling both customization and conservation of resources. While these methods typically tackle individual challenges, a comprehensive approach that can address all the aforementioned challenges remains yet to be thoroughly explored in research.

In this paper, we aim to address a pivotal question: \textit{How do we leverage FL to exploit a compact global representation under data heterogeneity that not only substantially reduces communication bandwidth and memory footprint but effectively enhances learning performance?} To this end, we propose a novel framework, namely \emph{\underline{Fe}derated \underline{D}eep \underline{EQ}uilibrium Learning} (\OurAlg), that leverages the alternating directions method of multipliers (ADMM)~\citep{boydadmm} consensus optimization to learn a compact, yet efficient global representation via deep equilibrium models (DEQ), combined with personalized explicit layers for each client. Such an optimization scheme can globally capture diverse patterns across clients' data while achieving local personalization. 

The main contributions of this work are summarized as follows.

\begin{itemize}[leftmargin=15pt]
	\item We introduce \OurAlg, a novel framework for \textit{resource-efficient} and \textit{personalized} FL under data heterogeneity. Central to \OurAlg is its ability to learn a \textit{compact global representation} with a low memory footprint via equilibrium models, which can be adapted to swiftly fine-tune \textit{personalized layers} at individual clients to facilitate personalization.
 
    \item To jointly learn global and personalized layers across clients' data, we formulate a consensus optimization problem rooted in ADMM. This problem is tactically decomposed into the primal problem -- seeking optimal personalized and shared parameters, and the dual problem -- controlling discrepancies between the local and global representation to mitigate client drift.
    
    \item We provide theoretical convergence analysis of \OurAlg for smooth nonconvex objectives. Under standard assumptions, we show that \OurAlg converges to a stationary point, concurrently achieving a global representation across all clients and optimal personalized models for each client.
	\item Rigorous experiments demonstrate that \OurAlg achieves performance comparable to state-of-the-art (SOTA) approaches while reducing communication demands by 2--4 times and memory footprint by 1.5 times. Additionally, the global representations prove effective in generalizing to unseen clients.
\end{itemize}

%% file: CIKMSections/related_works.tex
\section{Related Work}
\label{sec:related_work}
\textbf{Federated learning (FL)} has witnessed significant growth and innovation~\citep{flchallenge} since the establishment of the de-facto FedAvg algorithm~\citep{fedavg}. However, deploying FL in an resource-constrained environment still confronts primary challenges of communication and memory limitations and data heterogeneity~\citep{flchallenge}. 
Prior works have proposed communication and computation-efficient strategies such as quantization~\citep{konecny_federated_2017,suresh_distributed_2017, pmlr-v108-reisizadeh20a} -- quantizing the precision of model weights to streamline local computations and minimize data transfer size, model compression~\citep{com1} or prunning~\citep{cikm_com} -- reducing model size to make it more manageable to transmit, and adaptive communication~\citep{comadaptive} -- dynamically adjusting the amount of data exchanged based on various factors. In addition, some ADMM-based frameworks have been developed to address this challenge~\citep{admm1, admm4}.
Meanwhile, \textit{data heterogeneity} refers to the non-identically distributed (non-i.i.d.) data distribution across clients~\citep{zhao_federated_2018}. This challenge have been tackled using techniques like data sharing \citep{li_convergence_2020}, data augmentation \citep{yoon2021fedmix, cikm_noniid} or variance reduction \citep{scaffold}, which ensure the globally aggregated model accurately reflects the underlying patterns of clients' data.

\textbf{Personalized Federated Learning (PFL)}  has recently gained significant attention to address data heterogeity \citep{Tan2021TowardsPF}. Regarding full model personalization, a simple yet effective personalization approach involves fine-tuning the global model's parameters for each client \citep{fedavgft, collins2022fedavg}. Various works adopted multi-task learning (MTL), enabling clients to learn their distinct data patterns while benefiting knowledge shared by others \citep{smith_federated_2018, Marfoq2021, ditto}. Meta-learning frameworks are also employed to develop an effective initial model to rapidly adapt to new heterogeneous tasks \citep{perfedavg, pfedme}, while model interpolation techniques are used to personalize models using a mixture of global and local models \citep{Hanzely2020FederatedLO, apfl}. Recent years, partial model personalization approaches \citep{fedper, hanzaley, singhal2021federated, fedrep, lin2022personalized, knnper, pfedgp, fedalt, cikm_pfl} have brought up the idea of exploiting the common representation in heterogeneous settings to learn a personalized model tailored for each device, enabling simple formulation and layer-wise flexibility; however, their applicability is limited to specific settings such as  linear representations \citep{fedrep} or pre-trained representations \citep{knnper}. 

\textbf{Deep Equilibrium Learning} has emerged as a prominent implicit deep learning model in recent years~\cite{implicitdl}. In contrast to the conventional layer stacking models, deep equilibrium models (DEQ) implicitly define layers through an equilibrium point (a.k.a. fixed-point) of an infinite sequence of computation~\citep{deq}. A general DEQ consists of a single implicit layer $f_\theta$ modeled as a fixed-point system $z^\star = f_\theta(z^\star; x)$. Here, $\theta$ is the model parameter, x is the input data, and the equilibrium $z^\star$ is approximated by fixed-point iterations or root-finding. Several works delved into the theoretical underpinnings of DEQ~\citep{implicitdl, kawaguchi2021on}, providing valuable insights into the stability and approximation properties of implicit layers. Recent works have successfully applied DEQ on large-scale tasks such as sequential modeling, and semantic segmentation \citep{deq, mdeq}. 
In FL, DEQ showcases as a promising model architecture for enhancing communication and memory efficiency~\cite{fldeq}. However, fully exploiting this approach for tackling data heterogeneity is still uncharted.

%% file: CIKMSections/problem_formulation.tex
\begin{figure*}[t]
	\centering
	\begin{subfigure}{0.34\linewidth}
		\centering
		\includegraphics[width=\linewidth]{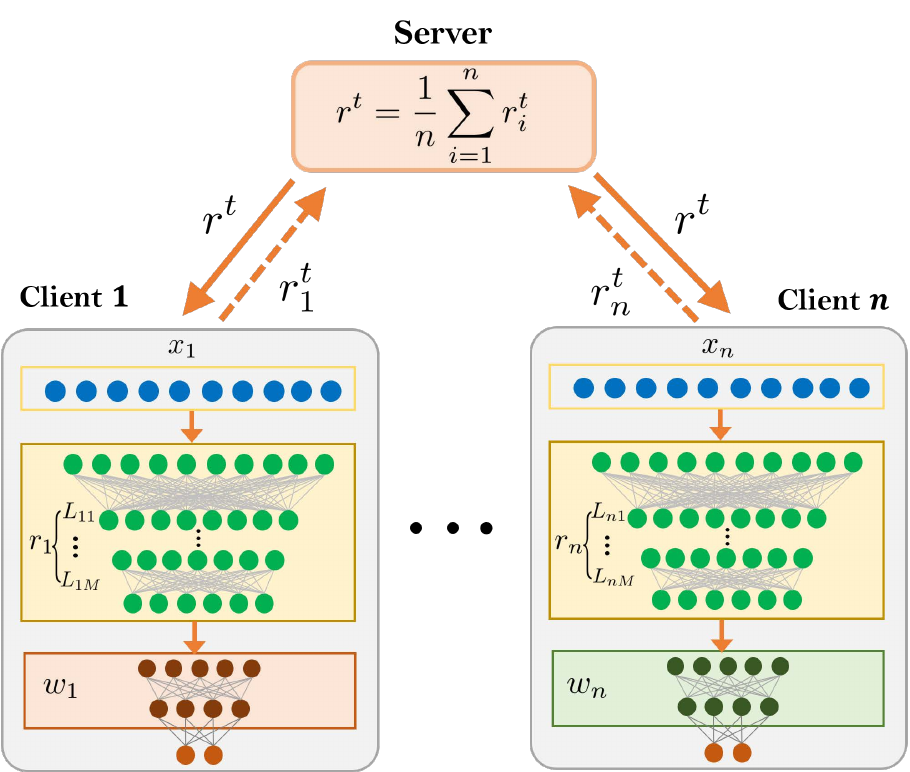}
		\caption{\label{fig:fedrep}} 
        \vspace{-5pt}
	\end{subfigure}
    \hspace{20pt}
	\begin{subfigure}{0.345\linewidth}
		\centering
		\includegraphics[width=\linewidth]{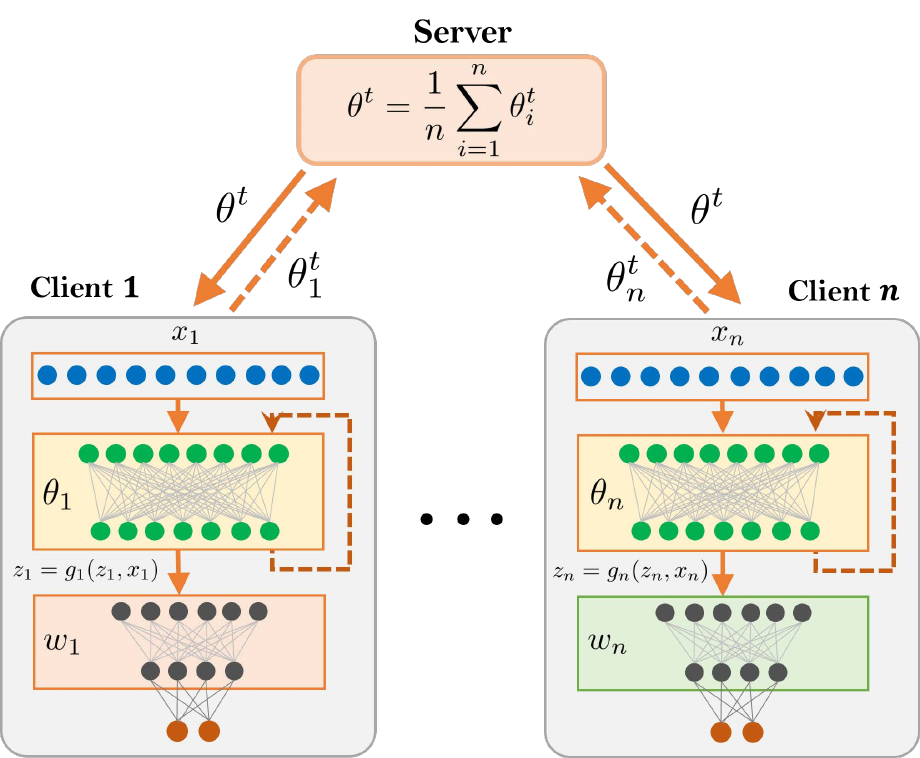}
		\caption{\label{fig:fedeq}} 
        \vspace{-5pt}
	\end{subfigure}
    \caption{The use of explicit and implicit layers as a global representation module for clients in FL. (a) Each client has $M$ explicit layers in its model. (b) Instead of having multiple explicit layers, each client uses only one equilibrium layer parametrized by $\theta_i$.}
    \label{F:fedeq_overview}
    \vspace{-5pt}
\end{figure*}

\section{Problem Formulation}
\label{sec:prob_form}
\subsection{Federated Deep Equilibrium Learning}
Consider a FL system comprising an aggregation server and $n$ clients. Each client $i$ owns a dataset of $n_i$ i.i.d. observations, denoted as $\{(x_{ij}, y_{ij}) \in \gX_i \times \gY_i : j = 1, \ldots, n_i\}$, capturing patterns and characteristics inherent to their domain. To distill meaningful insights from this data, we introduce a predictive function $f_i: \gX_i \rightarrow \gY_i$, tailored to the unique statistical properties of client's data, and a loss function $\ell(f_i(x_{ij}), y_{ij})$ measuring the discrepancy between $f_i(x_{ij})$ and $y_{ij}$. The local empirical risk on client $i$, defined as $\gL_i  = \frac{1}{n_i} \sum_{j=1}^{n_i} \ell(f_i(x_{ij}), y_{ij})$, reflects how well the model is performing within the individual local context. The primary objective is to minimize the sum of all local risks. Conventional FL approaches, e.g., FedAvg~\citep{fedavg}, learn a shared model by averaging clients' parameters after local stochastic gradient descent (SGD) rounds. However, this method does not account for data heterogeneity, which can result in \textit{client drift}.  
One way to address this is partial personalization, where a model is decoupled into a \emph{representation module} and a \emph{personalized module}. Clients collaboratively learn a common (linear or non-linear) representation to capture diverse patterns of all data, while further adapting their personalized modules for better local generalization \citep{fedper, fedrep, knnper}.
An effective representation, however, is often deep with many parameters. For example, the representation $r$ illustrated in \Cref{fig:fedrep} consists of $M$ layers implying a high communication cost and $O(M)$ memory complexity during training. These costs can prove to be significant barriers, particularly in resource-constrained environments where efficiency is paramount.
In this work, we concurrently tackle the aforementioned challenges through the utilization of deep equilibrium learning.
Specifically, for every client $i$, we consider a special predictive function $f_i$ of the following form: 
\begin{equation}
 	\begin{aligned}
 	f_{\theta_i, w_i}\left(x_{ij}\right) & \defeq h_{w_i}\left(g_{\theta_i}\left(z_{i j}^{\star}; x_{i j}\right)\right) \\
 	\text{s.t.} \quad z_{i j}^{\star} & = g_{\theta_i}\left(z_{i j}^{\star}; x_{i j}\right),  \quad \forall i, j
 	\end{aligned}
 	\label{F:opt_f}
\end{equation}
By this definition, $f_i$ is a composite of two functions: $g$ and $h$ parametrized by $\theta_i$ and $w_i$, respectively. While $h_{w_i}$ can be any conventional layers, such as a fully-connected network, $g_{\theta_i}$ is defined implicitly with the constraint $z_{i j}^{\star} = g_{\theta_i}(z_{i j}^{\star}; x_{i j})$, as depicted in~\Cref{fig:fedeq}. Such a representation for $g_{\theta_i}$ is called a \emph{deep equilibrium model}~\citep{deq} with key advantages of compactness and expressive power: a single layer can represent an infinitely deep weight-tied network while maintaining a constant memory footprint~\citep{deq}. This allows the model to learn a sophisticated non-linear representation at a low memory capacity. We describe this in detail in~\Cref{sec:prob_form:deq_layer}.\

We then formulate a global consensus optimization problem for \OurAlg as follows.
\begin{equation}
 	\begin{aligned}
 		\min_{\{\theta_i\}, \{w_i\}, \theta}
 		\quad \sum_{i=1}^n\Bigg[\gL_i\left(\theta_i, w_i\right) &  \defeq \frac{1}{n_i} \sum_{j=1}^{n_i} \ell \left(f_{\theta_i, w_i}\left(x_{i j}\right), y_{i j}\right)\Bigg] \\
 		\text{s.t.}
 		\quad \theta_i & = \theta,  \quad \forall i
 	\end{aligned}
 	\label{F:opt_problem}
 \end{equation}

The aim of \Cref{F:opt_problem} is to minimize the global loss under the consensus constraints $\theta_i = \theta, \forall i$ to enforce that the local $\theta_i$ are learned through sharing among clients. This encourages the representation to benefit from the structure of all clients' data. Meanwhile, all $w_i$ are fine-tuned locally, adapting from the shared global representation to achieve personalization. 

\subsection{Equilibrium Representation}
\label{sec:prob_form:deq_layer}

\input{./CIKMSections/deq}

\section{\OurAlg: Algorithm Design}
\label{sec:consensus_opt}

\subsection{Learning Global Representation via ADMM Consensus Optimization}

With the objective \Cref{F:opt_problem}, we propose to employ an ADMM-based~\citep{boydadmm} framework to jointly learn a global equilibrium representation and personalized layers across clients.
This approach is based on alternating updates of primal and dual variables to achieve two goals: (1) $\theta_i$ across clients become closer and (2) the objective function decreases. To fulfill this, we first introduce a dual variable $\lambda_i$ for each $\theta_i$ and a penalty $\rho > 0$, then construct the augmented Lagrangian for each client $i$ as follows.
\begin{equation}
	\label{eq:prob_form:admm:augmented_langrangian}
    \tilde{\gL_i}(\theta, \theta_i, w_i, \lambda_i) \defeq \gL_i(\theta_i, w_i) + \innProd{\lambda_i, \theta_i - \theta} + \frac{\rho}{2} \norm{\theta_i - \theta}^2.
\end{equation}

Upon solving~\Cref{eq:prob_form:admm:augmented_langrangian}, clients will then collaboratively minimize the global objective $\widetilde{\gL}$, which is defined as follows.
\begin{equation}
	\label{eq:prob_form:admm:augmented_langrangian_g}
    \widetilde{\gL}(\theta, \{ \theta_i \}, \{ w_i \}, \{ \lambda_i \}) \defeq \sum_{i=1}^{n} \left[ \tilde{\gL_i}(\theta, \theta_i, w_i, \lambda_i) \right].
\end{equation}

\textbf{Combating Client Drift through ADMM Consensus:} Delving deeper into equation~\Cref{eq:prob_form:admm:augmented_langrangian}, two essential terms come to the forefront. The inner product term, $\innProd{\lambda_i, \theta_i - \theta}$, reflects the alignment between the deviation of a local model from the global model and its associated dual variable, $\lambda_i$. Essentially, it acts as a controller of discrepancies between local and global models. The quadratic term, $\frac{\rho}{2} \norm{\theta_i - \theta}^2$, directly penalizes deviations of the local model from the global standard. The parameter $\rho$ adjusts the stringency of this penalty, ensuring a balance between local adaptability and global consensus. These terms result in the new correction term in the gradient $\nabla_{\theta_i} \tilde{\gL_i} = \nabla_{\theta_i}\gL_i + \lambda_i + \rho (\theta_i - \theta)$ aligning client updates with server directions to overcome drifting.

\begin{algorithm}[t]
	\caption{Federated Deep Equilibrium Learning (\OurAlg)}\label{alg:fedeq_alg}
    \small
	\begin{algorithmic}[1]
		\STATE \textbf{Parameters:} $T$ communication rounds; penalty $\rho$
		\STATE Initialize $S_0 = \{1, \ldots, n\}$, and $\theta^0 = \theta_i^0$, $w_i^0$ and $\lambda_i^0$ randomly $\forall i$
		\FOR{$t = 0, 1, \ldots$, $T-1$} 
		\STATE Server updates the shared parameters: $\theta^{t+1} = \frac{1}{\abs{S_{t}}} \sum_{i \in S_{t}} \theta_i^{t}$
		\STATE Server randomly samples a subset $S_{t+1}$ of clients
		\FOR{each client $i \in S_{t+1}$ in parallel}
		\STATE Receive $\theta^{t+1}$ from server 
		\\ Update primal variables:
		\STATE \hspace{0em}\raisebox{-0.7\baselineskip}[0pt][0pt]{$ \left\{\rule{0pt}{1.2\baselineskip}\right.$} 
		$w_i^{t+1}\leftarrow \argmin_{w_i} \gL_i(\theta^{t+1}, w_i)$ 
		\STATE \hspace{1em}$\theta_i^{t+1} \leftarrow \textsc{Rep\_Update}(\tilde{\gL}_i(\theta^{t+1}, \theta_i^{t}, w_i^{t+1}, \lambda_i^{t}))$ 
		
		\STATE Update dual variable: $\lambda_i^{t+1} = \lambda_i^{t} + \rho \left( \theta_i^{t+1} - \theta^{t+1} \right)$
		\STATE Send $\theta_i^{t+1}$ to server 
		\ENDFOR
		\ENDFOR
		\ALGPHASE{}
		\textbf{function} \textsc{Rep\_Update}($\tilde{\gL_i}(\theta, \theta_i, w_i, \lambda_i$)
		\STATE \textbf{Parameters:} local dataset ($\gX_i, \gY_i$), learning rate $\eta$
		\FOR{each local epoch}
		\FOR{each sample $(x,y) \in (\gX_i, \gY_i)$}
		\STATE Find the fixed-point $z^{\star} = g_{\theta_i}(z^{\star}; x)$ using Anderson Acceleration with warm-start
		\STATE Compute the loss $\gL_i(h_{w_i}(g_{\theta_i}(z^{\star}; x)), y)$ and the gradient $\nabla_{\theta_i} \gL_i$ using Eq. \Cref{F:ift}
		\STATE Update gradients $\nabla_{\theta_i} \tilde{\gL_i} = \nabla_{\theta_i}\gL_i + \lambda_i + \rho (\theta_i - \theta)$
		\STATE Update: $\theta_i \leftarrow \theta_i - \eta \nabla_{\theta_i} \tilde{\gL_i}$
		\ENDFOR
		\ENDFOR
		\STATE \textbf{return} $\theta_i$
	\end{algorithmic}
\end{algorithm}

As described in~\Cref{alg:fedeq_alg}, the ADMM consensus proceeds by iteratively updating the local parameters $\theta_i$ and $w_i$, the dual variables $\lambda_i$, and the shared global parameters $\theta$. This procedure consists of three key steps per round of communication. 
First, the server updates the shared parameters $\theta$ based on the set of $\theta_i$ obtained from the previous round (line 4). It then distributes these updated parameters to a new subset of clients (line 5 and line 7). Second, each selected client $i$ minimizes the augmented Lagrangian $\tilde{\gL}_i$ w.r.t. its primal variables (lines 8--9). This minimization is performed using the \textsc{Rep\_Update} procedure (lines 12--19), which involves solving problems \Cref{F:opt_f} and \Cref{F:ift} to obtain the gradient $\nabla_{\theta_i} \tilde{\gL_i}$ via implicit differentiation. The dual variable $\lambda_i$ is then updated by an ascent step to enforce the consensus constraint (lines 10). This process is repeated for several iterations until convergence, at which point the local variable $\theta_i$ and the consensus variable $\theta$ agree.

\subsection{Personalization through Explicit Layers}

With the global representation learned through ADMM consensus optimization, we proceed to the second aim of \OurAlg, which focuses on personalizing the model for each client to better address data heterogeneity. Particularly, in every communication round, client $i \in S$ fine-tunes its personalized parameters $w_i$ on top of the global representation $\theta$ with its local data, i.e., minimizing the original loss $\gL_i(\theta, w_i)$ w.r.t. $w_i$ (line 8 of \Cref{alg:fedeq_alg}). This problem can be efficiently solved by using gradient-based optimization methods such as SGD, where the personalized parameter $w_i^{t}$ obtained from the previous round $t$ is used as the initial value for updating the next $w_i^{t+1}$, providing clients the flexibility adapt to their unique data distributions to different personalization settings.

Combining equilibrium learning with the ADMM consensus optimization framework enhances FL in key ways. Equilibrium models, with their implicitly "infinite-depth" single layer, \textit{simplify model complexity} and \textit{reduce communication overhead} by requiring fewer parameters. Moreover, they maintain \textit{a constant, often lower, memory footprint} than traditional deep models. Additionally, the ADMM framework permits \textit{asynchronous updates of local parameters}, thereby optimizing the utilization of network resources and the convergence speed.

\subsection{Convergence Analysis}

We theoretically analyze \OurAlg's convergence in a smooth, non-convex setting. We note three differences from the vanilla ADMM \cite{boydadmm} that are in \Cref{alg:fedeq_alg}. First, only a subset of clients are chosen to participate in each round (line 5 of \Cref{alg:fedeq_alg}), a common setting in FL. Second, clients' parameters are split into a shared module ($\theta_i$) and a local-only module ($w_i$). Third, local optimization (lines 8--9 of \Cref{alg:fedeq_alg}) is performed approximately using a sequence of SGD updates.

\begin{assumption} 
    For every client $i$, the local loss function $\gL_i$ is bounded below. Further, $\gL_i$ is differentiable everywhere w.r.t. $(\theta_i, w_i)$ and has Lipchitz gradients with parameter $L > 0$.
    \label{assumption:admm}
\end{assumption}

To establish the convergence of \OurAlg in a non-convex setting of ADMM, we aim to show the following. Start with the variables $(\theta^t, \{ \theta_i^t \}, \{ w_i^t \}, \{ \lambda_i^t \})$. After $T$ rounds (i.e., after all clients have participated at least once since iteration $t$), the new variables
$$(\theta^{t+T}, \{ \theta_i^{t+T} \}, \{ w_i^{t+T} \}, \{ \lambda_i^{t+T} \})$$ lead to decrease in the global augmented Lagrangian $\widetilde{\gL}$. Then we show that these variables will converge to a local stationary point of $\widetilde{\gL}$.

\begin{lemma}[Decrease in augmented Lagrangian after every $T$ iterations]
    Suppose $\rho$ is chosen large enough that $\rho > 6L$. After $T$ global iterations since $t$, when all clients have participated in training at least once, we have
    \begin{align}
        & \widetilde{\gL}(\theta^{t+T}, \{ \theta_i^{t+T} \}, \{ w_i^{t+T} \}, \{ \lambda_i^{t+T} \}) - \widetilde{\gL}(\theta_{t}, \{ \theta_i^{t} \}, \{ w_i^{t} \}, \{ \lambda_i^{t} \}) \nonumber \\
        & \leq \frac{1}{T} \sum_{i=1}^n \left[
            - \left( \frac{\rho}{2} - \frac{3 L^2}{\rho} \right) \norm{\theta^{t+T} - \theta^{t}}^2 \right.     \label{eq:auglag_decrease} \\
            & ~ \qquad \left. - \left( \frac{L + \rho}{2} - \frac{12 L^2}{\rho} \right) \norm{\theta_i^{t+T} - \theta_i^{t}}^2 - \left( \frac{L}{2} - \frac{3 L^2}{\rho} \right) \norm{w_i^{t+T} - w_i^{t}}^2 \right]. \nonumber
    \end{align}
    \label{thrm_proof:lemma:auglag_decrease}
\end{lemma}

\begin{proof}
First, the change in the augmented Lagrangian after one global round can be decomposed as follows
\begin{align}
    & \widetilde{\gL}(\theta^{t+1}, \{ \theta_i^{t+1} \}, \{ w_i^{t+1} \}, \{ \lambda_i^{t+1} \}) - \widetilde{\gL}(\theta^t, \{ \theta_i^t \}, \{ w_i^t \}, \{ \lambda_i^t \}) \nonumber \\
    & = \quad \left( \widetilde{\gL}(\theta^{t+1}, \{ \theta_i^{t} \}, \{ w_i^{t} \}, \{ \lambda_i^{t} \}) - \widetilde{\gL}(\theta^t, \{ \theta_i^t \}, \{ w_i^t \}, \{ \lambda_i^t \}) \right) \label{eq:admm:auglag_split_1} \\
    & \quad + \left( \widetilde{\gL}(\theta^{t+1}, \{ \theta_i^{t+1} \}, \{ w_i^{t+1} \}, \{ \lambda_i^{t} \}) - \widetilde{\gL}(\theta^{t + 1}, \{ \theta_i^t \}, \{ w_i^t \}, \{ \lambda_i^t \}) \right) \nonumber\\
    & \quad + \left( \widetilde{\gL}(\theta^{t+1}, \{ \theta_i^{t+1} \}, \{ w_i^{t+1} \}, \{ \lambda_i^{t+1} \}) - \widetilde{\gL}(\theta^{t+1}, \{ \theta_i^{t+1} \}, \{ w_i^{t+1} \}, \{ \lambda_i^t \}) \right). \nonumber
\end{align}
Each of the three differences on the right-hand side corresponds to the sequential update rule for the variables in \Cref{alg:fedeq_alg}: $\theta$, then $\theta_i$ and $w_i$, and finally $\lambda_i$.

\textbf{Bounding the first term of \Cref{eq:admm:auglag_split_1}.} Note that by the definitions in \Cref{eq:prob_form:admm:augmented_langrangian,eq:prob_form:admm:augmented_langrangian_g}, $\widetilde{\gL}_i$ is $(n\rho)$-strongly convex w.r.t. $\theta$. The averaging step to obtain $\theta^{t+1}$ (line 4 of \Cref{alg:fedeq_alg}) is basically performing the minimization of $\widetilde{\gL}$ w.r.t. $\theta$. In other words, $$\theta^t = \argmin_{\theta} \widetilde{\gL}(\theta, \{ \theta_i^t \}, \{ w_i^t \}, \{ \lambda_i^t \}).$$ 
Therefore, 
\begin{align}
    &\widetilde{\gL}(\theta^{t+1}, \{ \theta_i^{t} \}, \{ w_i^{t} \}, \{ \lambda_i^{t} \}) - \widetilde{\gL}(\theta^t, \{ \theta_i^t \}, \{ w_i^t \}, \{ \lambda_i^t \}) \nonumber\\
    & \qquad \qquad \leq - \frac{n \rho}{2} \norm{\theta^{t+1} - \theta^t}^2.
    \label{eq:admm:delta_1}
\end{align}

\textbf{Bounding the second term of \Cref{eq:admm:auglag_split_1}.} The second difference quantifies the changes in the augmented Lagrangian after updating the local variables $\theta_i$ and $w_i$. Because we perform client sampling, for those clients who are not chosen at each round, their augmented Lagrangian does not change: 
\begin{align}
    w_i^{t+1} = w_i \quad \text{and} \quad \theta_i^{t+1} = \theta_i^t, \quad \forall i \notin S_{t+1}.
    \label{eq:admm:delta_2:notchosen}
\end{align}

For the chosen client, analyze the change as follows. Since $\gL_i$ is assumed to be $L$-smooth w.r.t. $w_i$, we have
\begin{align*}
    \gL_i(\theta_i^t, w_i) \leq \gL_i(\theta_i^t, w_i^t) + \innProd{\nabla_{w_i} \gL_i(\theta_i^t, w_i^t), w_i - w_i^t} + \frac{L}{2} \norm{w_i - w_i^t}^2.
\end{align*}
By minimizing the right-hand side, we have the GD rule: $w_i^{t+1} = w_i^t - \frac{1}{L} \nabla_{w_i} \gL(\theta_i^t, w_i^t).$ This leads to the following sufficient decrease to the local loss
\begin{align*}
    \gL_i(\theta_i^t, w_i^t) - \gL_i(\theta_i^t, w_i^{t}) \leq - \frac{L}{2} \norm{w_i^{t+1} - w_i^t}^2.
\end{align*}
Since the augmented Lagrangian only differs from the local loss function by a constant w.r.t. $w_i$, this implies that
\begin{align}
    \widetilde{\gL}_i(\theta^{t+1}, \theta_i^t, w_i^{t+1}, \lambda_i^t) - \widetilde{\gL}_i(\theta^{t+1}, \theta_i^t, w_i^{t}, \lambda_i^t) \leq - \frac{L}{2} \norm{w_i^{t+1} - w_i^t}^2.
    \label{eq:admm:delta_2:w_i}
\end{align}

To bound the difference but w.r.t. $\theta_i$, note that as per the definition in \Cref{eq:prob_form:admm:augmented_langrangian}, $\widetilde{\gL}_i$ is $(L + \rho)$-smooth w.r.t. $\theta_i$. Using the same logic as above, this time updating $\theta_i$ using the gradient of the augmented Lagrangian, we have the following bound
\begin{align}
    \widetilde{\gL}_i(\theta^{t+1}, \theta_i^{t+1}, w_i^{t+1}, \lambda_i^t) - \widetilde{\gL}_i(\theta^{t+1}, \theta_i^{t}, w_i^{t+1}, \lambda_i^t) \leq - \frac{L + \rho}{2} \norm{\theta_i^{t+1} - \theta_i^t}^2.
    \label{eq:admm:delta_2:theta_i}
\end{align}

Combining \Cref{eq:admm:delta_2:notchosen,eq:admm:delta_2:w_i,eq:admm:delta_2:theta_i}, we have $\forall i$,
\begin{align}
    &\widetilde{\gL}_i(\theta^{t+1}, \theta_i^{t+1}, w_i^{t+1}, \lambda_i^t) - \widetilde{\gL}_i(\theta^{t+1}, \theta_i^t, w_i^{t}, \lambda_i^t) \nonumber \\
    & \leq - \frac{L}{2} \norm{w_i^{t+1} - w_i^t}^2 - \frac{L + \rho}{2} \norm{\theta_i^{t+1} - \theta_i^t}^2.
    \label{eq:admm:delta_2}
\end{align}

\textbf{Bounding the third term of \Cref{eq:admm:auglag_split_1}.} Based on the update of $\theta_i^{t+1}$ on line 10 of \Cref{alg:fedeq_alg}, the change in the augmented Lagrangian can be written as
\begin{align}
    &\widetilde{\gL}_i(\theta^{t+1}, \theta_i^{t+1}, w_i^{t+1}, \lambda_i^{t+1}) - \widetilde{\gL}_i(\theta^{t+1}, \theta_i^{t+1}, w_i^{t+1}, \lambda_i^t) \nonumber \\ 
    &= \innProd{\lambda_i^{t+1} - \lambda_i^{t}, \theta_i^{t+1} - \theta^{t+1}}  = \frac{1}{\rho} \norm{\lambda_i^{t+1} - \lambda_i^{t}}^2.
    \label{eq:admm:delta_3:auglag_diff_local}
\end{align}
Notice that the first-order condition for $\theta_i^{t+1}$ implies that
\begin{align*}
    \nabla_{\theta_{i}} \gL_i(\theta_i^{t+1}, w_i^{t+1}) + \lambda_i^t + (L + \rho) (\theta_i^{t+1} - \theta^{t+1}) = 0.
\end{align*}
Using the fact that $\lambda_i^{t+1} = \lambda_i^{t} + \rho (\theta_i^{t+1} - \theta_t)$, we have
\begin{align}
    \lambda_i^{t+1} = - \nabla_{\theta_{i}} \gL_i(\theta_i^{t+1}, w_i^{t+1}) - L (\theta_i^{t+1} - \theta^{t+1}).
    \label{eq:admm:delta_3:lambda}
\end{align}
Therefore,
\begin{align*}
    \norm{\lambda_i^{t+1} - \lambda_i^{t}} & = \left\| \left[\nabla_{\theta_{i}} \gL_i(\theta_i^{t+1}, w_i^{t+1} - \nabla_{\theta_{i}} \gL_i(\theta_i^{t}, w_i^{t})) \right] \right. \\
    & \qquad \qquad \qquad \qquad \left. + \ L \left[ \theta_i^{t+1} - \theta_i^{t} \right] + L \left[ \theta^t - \theta^{t+1} \right] \right\| \\
    & \quad \leq 2 \left\| \theta_i^{t+1} - \theta_i^t \right\| + \left\| w_i^{t+1} - w_i^{t} \right\| + \left\| \theta^{t+1} - \theta^t \right\|,
\end{align*}
where on the second row, we use the triangle inequality and the fact that $\gL_i$ is $L$-smooth w.r.t. $(\theta_i, w_i)$. This implies that 
\begin{align}
    \norm{\lambda_i^{t+1} - \lambda_i^{t}}^2 \leq 12 L^2 \left\| \theta_i^{t+1} - \theta_i^t \right\|^2 + 3 L^2 \left\| w_i^{t+1} - w_i^{t} \right\|^2 + 3 L^2 \left\| \theta^{t+1} - \theta^t \right\|^2.
    \label{eq:admm:delta_3:lambda_normsq}
\end{align}
This result also holds for clients who do not participate in round $t$, because the left-hand side is simply zero.

Substituting \Cref{eq:admm:delta_3:lambda_normsq} into the right-hand side of \eqref{eq:admm:delta_3:auglag_diff_local} gives
\begin{align}
    & \widetilde{\gL}_i(\theta^{t+1}, \theta_i^{t+1}, w_i^{t+1}, \lambda_i^{t+1}) - \widetilde{\gL}_i(\theta^{t+1}, \theta_i^{t+1}, w_i^{t+1}, \lambda_i^t) \label{eq:admm:delta_3} \\
    & \leq \frac{12 L^2}{\rho} \left\| \theta_i^{t+1} - \theta_i^{t} \right\|^2 + \frac{3 L^2}{\rho} \left\| w_i^{t+1} - w_i^{t} \right\|^2 + \frac{3 L^2}{\rho} \left\| \theta^{t+1} - \theta^{t} \right\|^2.
     \nonumber
\end{align}

\textbf{Combining the differences.} Adding together \Cref{eq:admm:delta_1,eq:admm:delta_2,eq:admm:delta_3}, summing the result over all clients and then the unrolling the result over $T$ global iteration gives
\begin{align}
\begin{split}
    & \widetilde{\gL}(\theta^{t+T}, \{ \theta_i^{t+T} \}, \{ w_i^{t+T} \}, \{ \lambda_i^{t+T} \}) - \widetilde{\gL}(\theta^{t}, \{ \theta_i^{t} \}, \{ w_i^{t} \}, \{ \lambda_i^{t} \}) \\
    & \leq \sum_{k=1}^{T} \sum_{i=1}^n \left[
        - \left( \frac{\rho}{2} - \frac{3 L^2}{\rho} \right) \norm{\theta^{t+k+1} - \theta^{t + k}}^2 \right. \\
        & \quad \quad \quad \quad \quad \quad \left. - \left( \frac{L + \rho}{2} - \frac{12 L^2}{\rho} \right) \norm{\theta_i^{t+k+1} - \theta_i^{t + k}}^2 \right.\\
        & \quad \quad \quad \quad \quad  \quad \quad\left. - \left( \frac{L}{2} - \frac{3 L^2}{\rho} \right) \norm{w_i^{t+k+1} - w_i^{t + k}}^2 \right].
\end{split}
\label{eq:admm:delta_split}
\end{align}
If we choose $\rho > 6L$, then $\left( \frac{\rho}{2} - \frac{3 L^2}{\rho} \right)$, $\left( \frac{L + \rho}{2} - \frac{12 L^2}{\rho} \right)$ and $\left( \frac{L}{2} - \frac{3 L^2}{\rho} \right)$ are all positive. This allows us to move the first sum into each squared norm, we have \Cref{eq:auglag_decrease}.
\end{proof}

\begin{remark}
    \Cref{thrm_proof:lemma:auglag_decrease} shows that the augmented Lagrangian decreases by a sufficient amount after all clients have participated at least once, if we choose a large enough $\rho$. In practice, however, the smoothness parameter is difficult to estimate. We, therefore, fine-tune $\rho$ from a predefined range of values.
\end{remark}


\begin{lemma}[Lower bound for augmented Lagrangian]
    Suppose that the global loss function $\gL(\theta, \{ w_i \})$ is lower bounded, i.e., $\underline{\gL} \defeq \min_{\theta, \{ w_i \}} \gL(\theta, \{ w_i \}) > - \infty$. If $\rho$ is chosen large enough that $\rho > 5L$, then the augmented Lagrangian is lower bounded by $\underline{\gL}$.
    \label{thrm_proof:lemma:auglag_bounded}
\end{lemma}

\begin{proof}
For each client $i \in S_{t+1}$, we have
\begin{align}
    & \widetilde{\gL}_i(\theta^{t+1}, \theta_i^{t+1}, w_i^{t+1}, \lambda_i^{t+1}) \nonumber \\  
    & = \gL_i(\theta_i^{t+1}, w_i^{t+1}) + \frac{\rho}{2} \norm{\theta_i^{t+1} - \theta^{t+1}}^2  + \innProd{\lambda_i^{t+1}, \theta_i^{t+1} - \theta^{t+1}} \nonumber \\
    & = \gL_i(\theta_i^{t+1}, w_i^{t+1}) + \frac{\rho}{2} \norm{\theta_i^{t+1} - \theta^{t+1}}^2 \nonumber \\
    & \quad - \innProd{\nabla_{\theta_{i}} \gL_i(\theta_i^{t+1}, w_i^{t+1}) + L (\theta_i^{t+1} - \theta^{t+1}), \theta_i^{t+1} - \theta^{t+1}} \nonumber \\
    & = \gL_i(\theta_i^{t+1}, w_i^{t+1}) + \frac{\rho - 2L}{2} \norm{\theta_i^{t+1} - \theta^{t+1}}^2 \nonumber\\
    & \quad + \innProd{\nabla_{\theta_{i}} \gL_i(\theta_i^{t+1}, w_i^{t+1}), \theta^{t+1} - \theta_i^{t+1}} ,
    \label{eq:admm:auglag_lowerbound_1}
\end{align}
where in the second inequality we substitute expression of $\lambda_i^{t+1}$ in \Cref{eq:admm:delta_3:lambda}. Now, since $\gL_i$ is $L$-smooth w.r.t. $\theta_i$, we have
\begin{align}
    & \gL_i(\theta^{t+1}, w_i^{t+1}) \nonumber \\
    & \quad \leq \gL_i(\theta_i^{t+1}, w_i^{t+1})  + \frac{L}{2} \norm{\theta_i^{t+1} - \theta^{t+1}}^2  \nonumber \\
        & \quad \quad + \innProd{\nabla_{\theta_{i}} \gL_i(\theta_i^{t+1}, w_i^{t+1}), \theta^{t+1} - \theta_i^{t+1}} \nonumber \\
    & \quad = \gL_i(\theta_i^{t+1}, w_i^{t+1}) + \frac{L}{2} \norm{\theta_i^{t+1} - \theta^{t+1}}^2 \nonumber\\
            & \quad \quad  + \innProd{\nabla_{\theta_{i}} \gL_i(\theta_i^{t+1}, w_i^{t+1}) - \nabla_{\theta_{i}} \gL_i(\theta^{t+1}, w_i^{t+1}), \theta^{t+1} - \theta_i^{t+1}} \nonumber \\
            & \quad \quad + \innProd{\nabla_{\theta_{i}} \gL_i(\theta^{t+1}, w_i^{t+1}), \theta^{t+1} - \theta_i^{t+1}}  \nonumber
\end{align}
\begin{align}
    \leq \gL_i(\theta_i^{t+1}, w_i^{t+1}) & + \frac{3L}{2} \norm{\theta_i^{t+1} - \theta^{t+1}}^2 \nonumber \\
    & \quad  + \innProd{\nabla_{\theta_{i}} \gL_i(\theta^{t+1}, w_i^{t+1}), \theta^{t+1} - \theta_i^{t+1}},
    \label{eq:admm:auglag_lowerbound_2}
\end{align}
where in the last line we use the Cauchy-Schwarz inequality and again the fact that $\gL_i$ is $L$-smooth. From \Cref{eq:admm:auglag_lowerbound_1,eq:admm:auglag_lowerbound_2}, we have
\begin{align}
    \widetilde{\gL}_i(\theta^{t+1}, \theta_i^{t+1}, w_i^{t+1}, \lambda_i^{t+1}) \geq \gL_i(\theta^{t+1}, w_i^{t+1}) + \frac{\rho - 5L}{2} \norm{\theta_i^{t+1} - \theta^{t+1}}^2. \nonumber
\end{align}
This implies that
\begin{align}
    & \widetilde{\gL}(\theta^{t+1}, \{ \theta_i^{t+1} \}, \{ w_i^{t+1} \}, \{ \lambda_i^{t+1} \}) \nonumber\\ 
    & \quad \geq \sum_{i=1}^{n} \left( \gL_i(\theta^{t+1}, w_i^{t+1}) + \frac{\rho - 5L}{2} \norm{\theta_i^{t+1} - \theta^{t+1}}^2 \right) \nonumber\\
    & \quad = \gL(\theta^{t+1}, \{ w_i^{t+1} \}) + \frac{\rho - 5L}{2} \sum_{i=1}^{n} \norm{\theta_i^{t+1} - \theta^{t+1}}^2.
\end{align}

Given the assumptions that (i) $\gL$ is be bounded below by $\underline{\gL}$ and (ii) $\rho > 5L$, we establish that $\widetilde{\gL}(\theta^{t+1}, \theta_i^{t+1}, w_i^{t+1}, \lambda_i^{t+1})$ is lower bounded by $\underline{\gL}$.
\end{proof}

Now we combine \Cref{thrm_proof:lemma:auglag_decrease,thrm_proof:lemma:auglag_bounded} for the following theorem.

\begin{theorem}[Convergence of the augmented Lagrangian]
    Suppose the following are true: (1) After $T$ rounds, all clients have participated in training at least once; (2) The global loss function $\gL(\theta, \{ w_i \})$ is bounded below by a finite quantity $\underline{\gL}$; and (3) The hyperparameter $\rho$ for the augmented Lagrangian is chosen such that $\rho > 6L$.
    \comment{
    \begin{itemize}[leftmargin=10pt]
        \item After $T$ rounds, all clients have participated in training at least once.
        \item The global loss function $\gL(\theta, \{ w_i \})$ is bounded below by a finite quantity $\underline{\gL}$.
        \item The hyperparameter $\rho$ for the augmented Lagrangian is chosen such that $\rho > 6L$.
    \end{itemize}
    }
    Then the augmented Lagrangian $\widetilde{\gL}$ will monotonically decrease and is convergent to a quantity of at least $\underline{\gL}$. Further, for all $i=1, \dots, n$ we have $\lim_{t \rightarrow \infty} \norm{\theta^{t+T} - \theta_i^{t+T}} = 0$.
    \label{appn:admm:auglag_convergence}
\end{theorem}

\begin{proof}
\Cref{thrm_proof:lemma:auglag_decrease} implies that the augmented Lagrangian decreases by a non-negative amount after every $T$ global iterations. Given \Cref{thrm_proof:lemma:auglag_bounded} and the fact that every client will be updated at least once in the interval $[t, t + T]$, we conclude that the limit $\lim_{t \rightarrow \infty} \widetilde{\gL}(\theta^{t+T}, \{ \theta_i^{t+T} \}, \{ w_i^{t+T} \}, \{ \lambda_i^{t+T} \})$ exists and is at least $\underline{\gL}$.

Now we prove the second statement. From \Cref{eq:admm:delta_split} and the fact that $\widetilde{\gL}$ converges, we conclude that as $t \rightarrow \infty$, 
\begin{align*}
    \norm{\theta^{t+T} - \theta^{t}} \rightarrow 0, \quad \norm{\theta_i^{t+T} - \theta_i^{t}} \rightarrow 0, \quad \norm{w_i^{t+T} - w_i^{t}} \rightarrow 0.
\end{align*}
Combining this with \Cref{eq:admm:delta_3:lambda_normsq}, we have $\norm{\lambda_i^{t+1} - \lambda_i^{t}} \rightarrow 0$. Based on the definition of $\lambda_i^{t+1}$, this implies that $\norm{\theta_i^{t+1} - \theta^{t+1}} \rightarrow 0$. 
\end{proof}


\Cref{appn:admm:auglag_convergence} establishes that the sequence of primal and dual variables updated after every $T$ global iterations of \Cref{alg:fedeq_alg} converges. Furthermore, we have $\norm{\theta_i^{t+1} - \theta^{t+1}} \rightarrow 0$, i.e., the consensus constraint is satisfied.
In the below theorem we present another guarantee that the limit of the sequence $(\theta^{t}, \{ \theta_i^{t} \}, \{ w_i^{t} \}, \{ \lambda_i^{t} \})$ is the a stationary solution for Problem \Cref{F:opt_problem}.

\begin{theorem}[\OurAlg's convergence]
    Suppose the assumptions in \Cref{appn:admm:auglag_convergence} hold. Then the limit point $(\theta^{*}, \{ \theta_i^{*} \}, \{ w_i^{*} \}, \{ \lambda_i^{*} \})$ of the sequence $(\theta^{t}, \{ \theta_i^{t} \},$ $\{ w_i^{t} \}, \{ \lambda_i^{t} \})$ is a stationary solution to Problem \Cref{F:opt_problem}. That is, for all $i$,
    \begin{align*}
        \nabla_{\theta_i} \gL_i(\theta^*, w_i^*) + \lambda_i^* = 0; \quad \nabla_{w_i} \gL_i(\theta^*, w_i^*) = 0; \quad \theta^* = \theta_i^*.
    \end{align*}
    \label{appn:admm:auglag_convergence_stationary}
\end{theorem}
\begin{proof}
    See the proof to \citep[Theorem 2.4]{admm_nonconvex}.
\end{proof}

%% file: CIKMSections/deq.tex
The equilibrium constraint in~\Cref{F:opt_f} embodies an ``infinitely deep'' network through a fixed-point system. Here we examine a common non-linear mapping of the form:
\begin{equation}
	\label{eq:prob_form:deq:forward}
	z = g_\theta(z; x) \defeq  \phi(Bz + Cx + b)
\end{equation}
where $x \in \RR^d$ is the input, $z \in \RR^{d_1}$ is the output, $\theta = (B \in \RR^{d_1 \times d_1}, C \in \RR^{d_1 \times d}, b \in \RR^{d_1})$ is the parameters and $\phi: \RR^{d_1} \rightarrow \RR^{d_1}$ is some nonlinear activation functions. Eq.~\Cref{eq:prob_form:deq:forward} implies that $z^{\star}$ is a fixed point of the function $z = g_\theta(z; x) $ for a given data point (or a mini-batch) $x$, equivalent to feeding the input through an infinite sequence of weight-tied layers of a feedforward network. Such a single layer, while much smaller in size, can perform on par with models containing many explicit layers~\citep{deq}. This extremely appeals to FL since communication and memory are often bottlenecks.

\subsubsection{Forward Pass -- Fixed-Point Iteration}

The fixed-point system in the general problem~\Cref{F:opt_f} or the special case~\Cref{eq:prob_form:deq:forward} allows an equilibrium layer to capture long-range dependencies as all parts of the input contribute to the equilibrium state $z^{\star}$. Moreover, in the process of finding $z^{\star}$, the model implicitly emphasizes features that are stable under $g_{\theta}$, promoting robust and consistent representations. 

\textbf{Convergence of Equilibrium Representation:}
The Banach fixed-point theorem assures convergence to a unique equilibrium representation if $g_{\theta}$ is a contraction mapping~\citep{banach}, i.e., a Lipschitz constant $L < 1$ ensures
$\|g_{\theta}(x) - g_{\theta}(y)\| \leq L \|x - y\|, \forall x, y$ in the domain of $g_{\theta}$. 
However, it's often challenging to confirm this criterion in practice, especially when $g_{\theta}$ represents complex, nonlinear functions in deep models.
Hence, we make two assumptions following the theory in~\citep{implicitdl}. First, $\phi$ is a component-wise non-expansive (CONE) mapping, i.e., the $k$th component of its output only depends on the $k$th component of its input and when operating on scalar input, $\phi$ is $1$-Lipschitz continuous: $\forall u, v \in \RR, \abs{\phi(u) - \phi(v)} \leq \abs{u - v}$. Many activation functions in deep learning, such as tanh, sigmoid, and (leaky) ReLU, satisfy this property. Second, the infinity norm of $B$, denoted by $\norm{B}_\infty$, is strictly bounded above by $1$. This promotes fixed-point convergence in certain iterative methods by limiting the size of the updates~\citep{implicitdl}. These together imply that $g_{\theta}$ in \Cref{eq:prob_form:deq:forward} is a contraction mapping, and a fixed point $z^{\star}$ exists for all $x$. 

\begin{figure}[!t]
    \centering
    \includegraphics[width=0.5\linewidth]{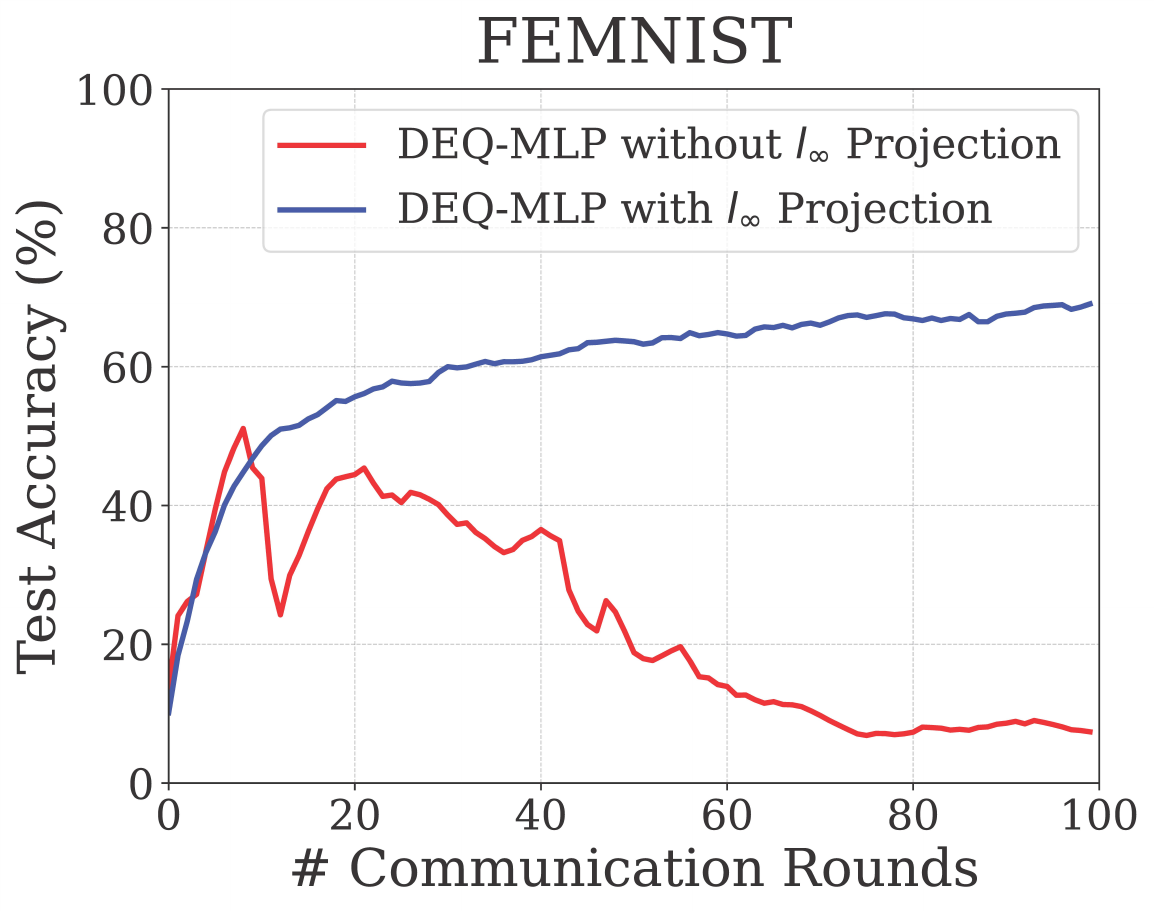}
    \caption{FL using DEQ-MLP with and without $l_\infty$ projection}
    \label{fig:deq_mlp}
\end{figure}

We demonstrate these findings to an FL context that involves 100 edge nodes, each possessing a portion of the FEMNIST dataset for a 62-class classification task. 
To benchmark, we employ the de-facto FedAvg algorithm and compare two local model configurations: 
(i) a DEQ-MLP model with a single equilibrium layer of size 512 (DEQ-MLP), wherein the parameters associated with $z$ are projected onto the $l_\infty$ norm, succeeded by a linear layer of size 128, and ultimately, a final layer mapping to 62 classes; and (ii) a similar architecture but without the projection step. As illustrated in \Cref{fig:deq_mlp}, DEQ-MLP with $l_\infty$ projection achieves convergence, whereas the model without projection appears to be diverging.

\textbf{Accelerating Fixed-point Solver:}
In this study, we employ a variant of Anderson Acceleration (AA)~\citep{aa} for solving fixed-point iterations in DEQ's forward pass~\citep{mdeq, ndeq}.
AA traditionally creates a new iterate by minimizing the discrepancies between previous iterates and their mappings over $m$ past iterations~\citep{aa}.
In terms of learning representation in \OurAlg, during training, the changes in $\theta$ and $x$ are small from one iteration to the next (i.e., $\theta^{(i)} \approx \theta^{(i-1)}$ and $x^{(i)} \approx x^{(i-1)}$), then the fixed points $z^{^{(i)}}$ and $z^{^{(i-1)}}$ are likely to be close to each other as well. Hence, we suggest a warm-start strategy for AA where the fixed point of the previous iterations is used as initialization for the next. This yields several benefits: (1) promotes \textit{faster convergence in fixed-point iterations} as the input data changes slightly from one iteration to the next; (2) enhances \textit{stability and robustness} against minor input or function changes by ensuring a smoother transition of learned representations across iterations; and (3) maintains \textit{context and continuity from previous data}, fortifying the model's resilience to such variations.

\subsubsection{Backward Pass -- Implicit Differentiation}

Similar to explicit models, DEQ's backward pass involves finding optimal model parameters to minimize a loss function using gradient-based methods (e.g., SGD). 
\comment{
Specially, to find the gradient of a loss $\gL$ w.r.t. $\theta$, one can employ the chain rule by following.
\begin{equation}
\frac{\partial \gL}{\partial \theta} = \frac{\partial \gL}{\partial z^{\star}} \frac{\partial z^{\star}(\theta)}{\partial \theta}
\label{eq:gradL_theta}
\end{equation}

In this context, the gradient of $\gL$ w.r.t. $z^\star$ can be derived through backpropagation. The notation $z^\star (\theta)$ is also utilized to signify situations where $z^\star$ is perceived as an implicit function of the variables that we're differentiating w.r.t. the shared parameter $\theta$, and we use $z^\star$ on its own when we're just mentioning the equilibrium value.}
However, due to the iterative nature of the fixed-point problem, deriving the gradient $\frac{\partial z^{\star}(\theta)}{\partial \theta}$ using conventional automatic differentiation is non-trivial. We, instead, can employ the implicit differentiation technique ~\citep{implicitft} as follows.

\begin{equation}
	\frac{\partial \gL}{\partial \theta} = \frac{\partial \gL}{\partial z^\star}\left ( I - \frac{\partial g(z^\star; x)}{\partial z^\star} \right )^{-1} \frac{\partial g(z^\star; x)}{\partial \theta },
	\label{F:ift}
\end{equation}
which is typically solved by using linear solvers like conjugate gradient descent~\citep{cjgd}. The main advantage of DEQ's backward pass is its memory efficiency. In explicit models, the gradients of intermediate variables need to be stored for backpropagation. In contrast, DEQ only need to store the final equilibrium state, resulting in substantial memory savings. It also allows for end-to-end training with gradient-based optimization methods~\citep{deq} as it is differentiable, and gradients can be computed and used for parameter updates.

\textbf{Reducing Complexity with Jacobian-Free Backpropagation:} The bottleneck of DEQ's backward pass lies in its high complexity. Recent studies show that the Jacobian inverse can be approximated using the zeroth-order approximation of the Neumann series~\citep{JFB} bypasses this costly computation, yet still provides good performance with significantly reducing training time. Thus, in this work, we employ this Jacobian-free backpropagation approach to estimate the implicit gradients for DEQ.

%% file: CIKMSections/experiments.tex
\section{Experimental Results}
\label{sec:experiment}

\subsection{Experimental Setup} 
\noindent
\par

\textbf{Datasets and Non-i.i.d. Partition:} We evaluate \OurAlg on four datasets: FEMNIST for handwritten character recognition~\citep{emnist}; CIFAR-10 and CIFAR-100 for image classification~\citep{cifar}; and Shakespeare for language modeling~\citep{fedavg}. We sample 200 clients for FEMNIST and Shakespeare from FedJax's datasets~\citep{fedjax}, following a natural partitioning method~\citep{caldas2019leaf}. The mean data size and standard deviation across clients' data are 195 $\pm$ 75.7 and 465 $\pm$ 506.2, respectively. For CIFAR-10/CIFAR-100, we employ conventional methods to sample heterogeneous local datasets~\citep{fedrep}, partitioning data equally among 100 clients which the number of labels per client is set at 2 and 5 for CIFAR-10, while CIFAR-100 has 5 and 20. 

\textbf{Baselines:} We compare \OurAlg against seven FL methods including: (1) locally trained models (Local); (2) FedAvg~\citep{fedavg}; (3) fine-tuning FedAvg (FedAvg+FT)~\citep{fedavgft, collins2022fedavg}; (4) Ditto~\citep{ditto}; (5) FedPer~\citep{fedper}; (6) FedRep~\citep{fedrep}; and (7) kNN-Per~\citep{knnper}, which the latter five are closely related to ours. FedAvg+FT and Ditto are well-known full model personalization approaches, while FedPer, FedRep and kNN-Per are emerging partial personalized FL methods enabling clients to learn an explicit shared representation and subsequently personalize the local model. We apply settings consistent with \OurAlg for all methods to ensure a fair comparison and fine-tune their respective hyperparameters to reach their best performance.

\textbf{Models and Representations:} For vision tasks, we construct DEQ-ResNet models inspired by ResNet \citep{resnet}, comprising an equilibrium residual block followed by a 3-layer fully connected (FC) network with the last serving for personalization and using Softplus activations \citep{softplus}, a smooth version of ReLU, to enforce the smoothness assumption in theoretical analysis. We offer two versions of DEQ-ResNet: "S" (small) and "M" (medium), varying by equilibrium layer's sizes. DEQ-ResNet-S and DEQ-ResNet-M are used for FEMNIST and CIFAR-10/CIFAR-100, respectively. As for baselines, we employ ResNet models, using their residual blocks with the same FC network as DEQ-ResNet. Especially, ResNet-20 and ResNet-34 are then employed for experiments on FEMNIST and CIFAR-10/CIFAR-100, correspondingly. For sequence tasks, we design a DEQ-Transformer based on Universal Transformer (UT)~\citep{uttransformer}, consisting of an equilibrium UT block succeeded by a personalized module with one last UT block and one linear layer; while other methods use a 8-layer, 4-head UT (Transformer-8). 

\textbf{Training Details:} We train models for $T= 150$ communication rounds in the experiments involving FEMNIST, CIFAR-10 and CIFAR-100, and $T=400$ rounds for Shakespeare. For all experiments, we uniformly sample 10\% of clients without replacement for each round and employ SGD optimizer with a constant learning rate. Unless specified in certain instances, we use $5$ local epochs for training global representation and $3$ for personalization. The final accuracy is averaged from local accuracies.


All experiments are conducted in a coding environment powered by an AMD Ryzen 3970X workstation with 256GB of RAM and fours NVIDIA GeForce RTX 3090 GPUs. For implementation, we ultilize advanced machine learning and optimization frameworks like Jax~\citep{jax}, Optax~\citep{optax}, Jaxopt~\citep{jaxopt}, FedJax~\citep{fedjax}, and Motley~\citep{motley}. 

\begin{figure*}[t]
	\centering
	\includegraphics[width=0.85\textwidth]{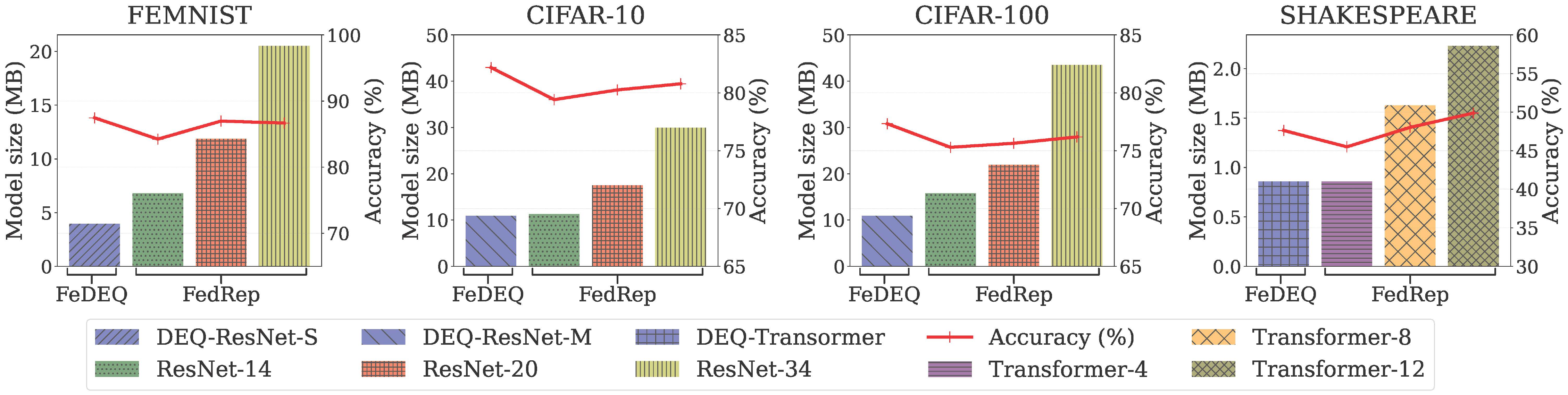}
	\caption{Comparative analysis of model size and test accuracy between \OurAlg and FedRep across various datasets.}
	\label{fig:model_size}
	\vspace{-5pt}
\end{figure*}

\subsection{Main Results}
\label{experiment:results}
\noindent
\par
\textbf{Model Size and Communication Efficiency:} We highlight the prowess of \OurAlg in achieving efficient communication, maintaining a compact model size without sacrificing performance. Specifically, we compare \OurAlg with DEQ against FedRep with explicit models. The results shown in Fig. \ref{fig:model_size} reveal that \OurAlg consistently maintains a smaller model size for all datasets while achieving competitive accuracy levels.
In FEMNIST, \OurAlg with DEQ-ResNet-S outperforms FedRep using the ResNet-14 model, despite being 1.5 times smaller in size. Remarkably, \OurAlg achieves performance metrics comparable to deeper architectures like ResNet-20 and ResNet-34, with a $2-4$ times smaller model size. This trend of maintaining compactness without compromising accuracy is also observed in CIFAR-10 and CIFAR-100, where \OurAlg either matches or slightly exceeds the performance of other benchmarked models. Additionally, in the Shakespeare dataset, \OurAlg's implementation with DEQ-Transformer consistently presents a model that is up to $3$ times more compact than standard Transformers, yet the accuracy remains competitive. 
A smaller model size implies more efficient update exchanges in FL scenarios, an essential aspect in FL environments characterized by limited bandwidth and resources.

\begin{table}[t]
	\centering
    
	\caption{Comparison of test accuracy on training clients between \OurAlg and other baselines across datasets. ($c$ and $p$ are the number of clients and data classes per client, respectively. For FEMNIST and Shakespeare, $p$ can vary denoted by "--")}
	\label{tab:per_acc}
    \scalebox{0.85}{
    \renewcommand{\arraystretch}{1.01}
    \begin{tabular}{@{}|c|cc|cc|cc|c|@{}}
        \hline
		\thead{\textbf{Datasets}} & \multicolumn{2}{c|}{\textbf{FEMNIST}} & \multicolumn{2}{c|}{\textbf{CIFAR-10}} & \multicolumn{2}{c|}{\textbf{CIFAR-100}} & \thead{\textbf{Shake} \\ \textbf{speare}} \\
        \hline
  		$\Big(\begin{matrix} c\\p \end{matrix}\Big)$ 
        & $\Big(\begin{matrix} 100\\ - \end{matrix}\Big)$
        & $\Big(\begin{matrix} 200\\ - \end{matrix}\Big)$ 
        & $\Big(\begin{matrix} 100\\ 2 \end{matrix}\Big)$ 
        & $\Big(\begin{matrix}100\\ 5 \end{matrix}\Big)$ 
        & $\Big(\begin{matrix} 100\\ 5 \end{matrix}\Big)$ 
        & $\Big(\begin{matrix} 100\\ 20 \end{matrix}\Big)$
        & $\Big(\begin{matrix} 200\\ - \end{matrix}\Big)$ \\
        \hline
        
		Local          & 68.80  & 68.77  & 83.85 & 65.78    & 66.67 & 34.93 & 28.70    \\ 
		FedAvg         & 83.50  & 82.12  & 38.91 & 63.78 & 33.05 & 33.06 & 48.07 \\ 
		FedAvg+FT    & 84.98  & 85.40  & 88.75 & 79.88 & 77.14 & 55.71 & 46.60 \\ 
        \hline
		Ditto          & \underline{87.93}  & \underline{88.43}  & 90.27 & 82.18 & 80.37 & 57.18    & 47.56   \\
		FedPer         & 85.10  & 85.42  & 87.89 & 80.38 & 78.88 & 55.07    & 44.66   \\ 
		FedRep         & 85.29  & 86.18  & 87.81 & 80.25 & 79.53 & 56.30 & 48.35   \\ 
		kNN-Per        & 84.94  & 85.36  & 87.43 &  80.18   & 76.19   & 56.27    & 48.46   \\ 
        \hline
		\textbf{\OurAlg}   & \textbf{87.13}  & \textbf{87.46} & \textbf{\underline{90.54}} & \textbf{\underline{82.84}} & \textbf{\underline{80.42}} & \textbf{\underline{57.49}} & \textbf{\underline{48.57}} \\ 
        \hline
	\end{tabular}
	}
 
	\vspace{-11pt}
\end{table}

\textbf{Personalization Performance:} 
We evaluate how \OurAlg addresses the data heterogeneity through personalization,  comparing its performance against various baselines on the local test datasets (unseen during training) across different tasks. As shown in Table.~\ref{tab:per_acc}, \OurAlg consistently outperforms non-personalized methods (Local, FedAvg, and FedAvg+FT), which generally exhibit lower performance under data heterogeneity. For instance, on the FEMNIST dataset, it achieved accuracies of 87.13\% and 87.46\% for 100 and 200 clients, respectively, outperforming the best non-personalized method by about 18\%.
When pitted against personalized FL baselines, \OurAlg showcases competitive performance with Ditto and faintly surpasses FedPer, FedRep, and kNN-Per. On FEMNIST, \OurAlg achieves accuracies that slightly lag behind Ditto but outshines other methods. 
In the CIFAR-10 dataset with various degrees of heterogeneity, \OurAlg achieves top accuracies, marginally surpassing those of other methods under the same conditions. Similarly, for the CIFAR-100 dataset, \OurAlg's performance peaks at 80.42\% and 57.49\% for configurations (100, 5) and (100, 20), respectively, on par with the closest competitor, Ditto.
Furthermore, in the next-character prediction task, \OurAlg achieves the highest accuracy, while kNN-Per closely follows with a small gap, exhibiting competitiveness. Consequently, \OurAlg consistently delivers comparable or superior performance with SOTA algorithms across all datasets while offering substantial efficiency in terms of communication with up to $4$ times smaller in the size of the representation. 

\begin{table}[t]
	\centering
	\caption{Comparison of test accuracy on unseen clients with test accuracy on training clients indicated in parentheses. ($c$, $u$, and $p$ are the number of training clients, unseen clients and data classes per client, respectively.)}
	\label{tab:unseen_acc}
     \renewcommand{\arraystretch}{1.01}
    \scalebox{0.85}{
	\begin{tabular}{@{}|c|c|c|c|c|@{}}
	   \hline
		 \textbf{Datasets}  & \textbf{FEMNIST} & \textbf{CIFAR-10} & \textbf{CIFAR-100} & \textbf{Shakespeare} \\
		\cline{2-5}
		  ($c$, $u$, $p$) & (180, 20, --)  &  (90, 10, 5) & (90, 10, 20) & (180, 20, --)   \\
		\hline
		FedAvg        & 80.04 (82.30)  & 50.20 (57.76)  & 28.40 (28.90)  & 46.92 (47.68)  \\ 
		Ditto    & 81.65 (88.70)   & 50.90 (82.34)   & 29.60 (55.86)  & 47.20 (47.24)  \\ 
		FedPer        & 83.47 (85.20)   & 78.20 (79.63)   & 59.20 (52.82)  & 45.47 (44.53)  \\ 
		FedRep        & 84.68 (86.12)  & 78.90 (79.11)   & 57.40 (52.97)  & 47.89 (48.00)   \\ 
		kNN-Per        & 84.48 (85.40)   & 75.70 (80.72)   & 50.70 (53.73)  & 48.26 (48.59)\\ 
        \hline
		\textbf{\OurAlg}  & \textbf{\underline{85.69}} (87.00)   & \textbf{\underline{79.20}} (83.18)  & \textbf{\underline{60.10}} (54.22)  & \textbf{\underline{48.35}} (48.56)  \\ 
        \hline
	\end{tabular}}
    \vspace{-10pt}
\end{table}
\textbf{Generalization to New (Unseen) Clients:}  To assess the quality of global representation in generalizing to unseen clients, we split the clients into two groups: 90\% training clients, and 10\% new, unseen clients. Initially, we train the shared representation for all methods using the designated training group, then use it to refine the personalized layers for new clients using their local data, which has not been incorporated during the representation training process. For FedAvg and Ditto, a full model will be trained and adapted to new clients. As shown in Table~\ref{tab:unseen_acc}, \OurAlg demonstrates remarkable competitiveness, consistently surpassing all other algorithms on FEMNIST, CIFAR-10, and CIFAR-100 while keeping comparable performance on Shakespeare. Although Ditto showcases effective personalization in training clients, it falls short when dealing with new clients. kNN-Per underperforms FedPer and FedRep in vision tasks as its shared representation is trained from scratch rather than using pre-trained models as in the original proposal, but excels in the sequence task.  This implies that the global representation facilitated by \OurAlg possesses the capacity for effective adaptation to newly joined clients, even with a significantly reduced size.

 \begin{figure}
	\centering
	\includegraphics[width=0.47\linewidth]{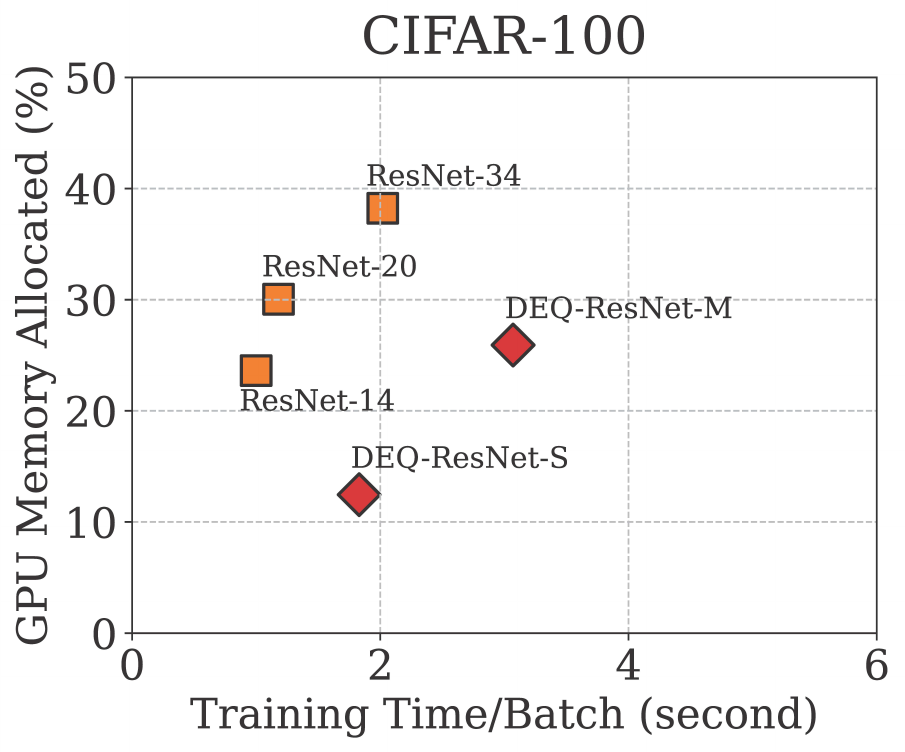}
    \vspace{-5pt}
	\caption{Comparison of memory usage and training time.}
	\label{fig:memory}
	\vspace{-5pt}
\end{figure}
\begin{figure}[t]
	\centering
	\begin{subfigure}{.47\linewidth}
		\centering
		\includegraphics[width=\linewidth]{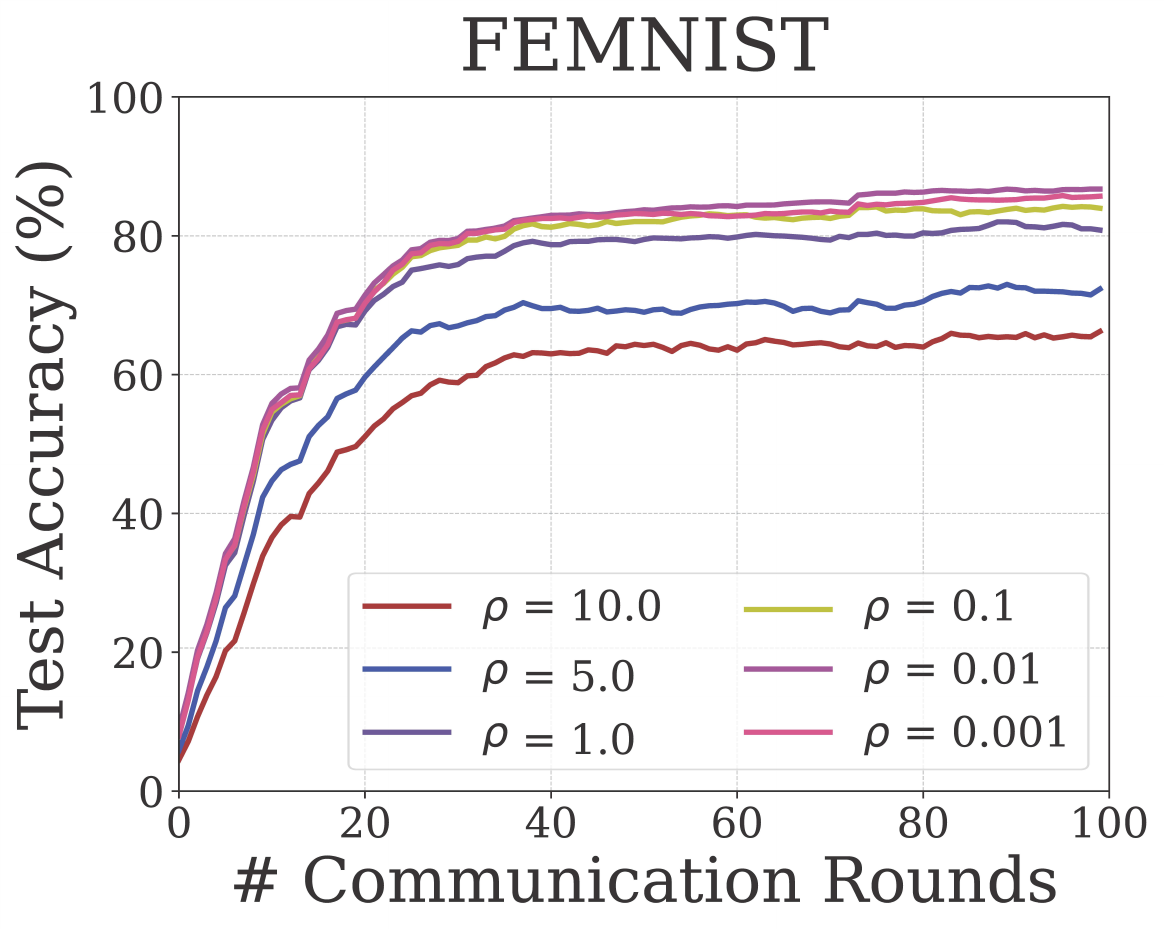}
	\end{subfigure}
	\begin{subfigure}{.47\linewidth}
		\centering
		\includegraphics[width=0.95\linewidth]{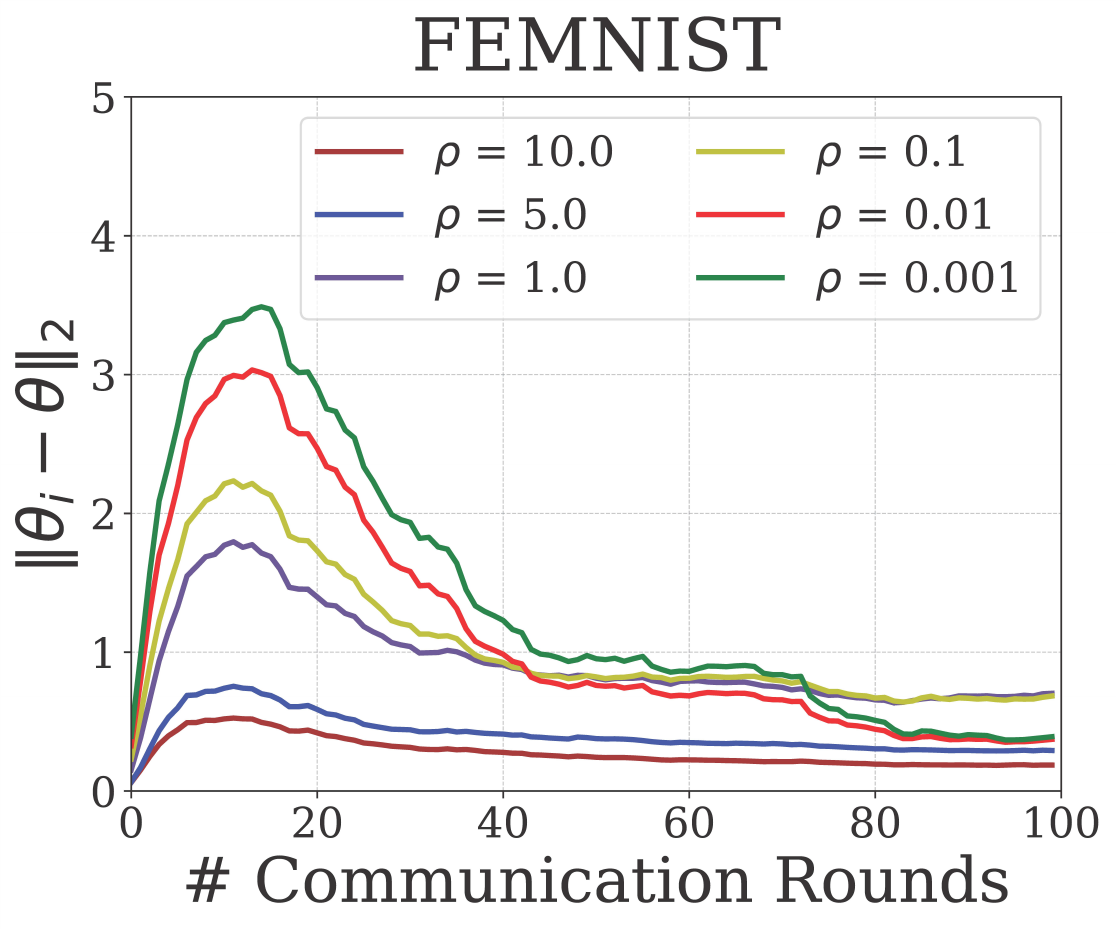}
	\end{subfigure}
    \vspace{-5pt}
	\caption{Effects of $\rho$ on \OurAlg's convergence.}
	\label{fig:rho_femnist}
    \vspace{-5pt}
\end{figure}

 \begin{figure}[t]
	\centering
	\begin{subfigure}{.47\linewidth}
		\centering
		\includegraphics[width=\linewidth]{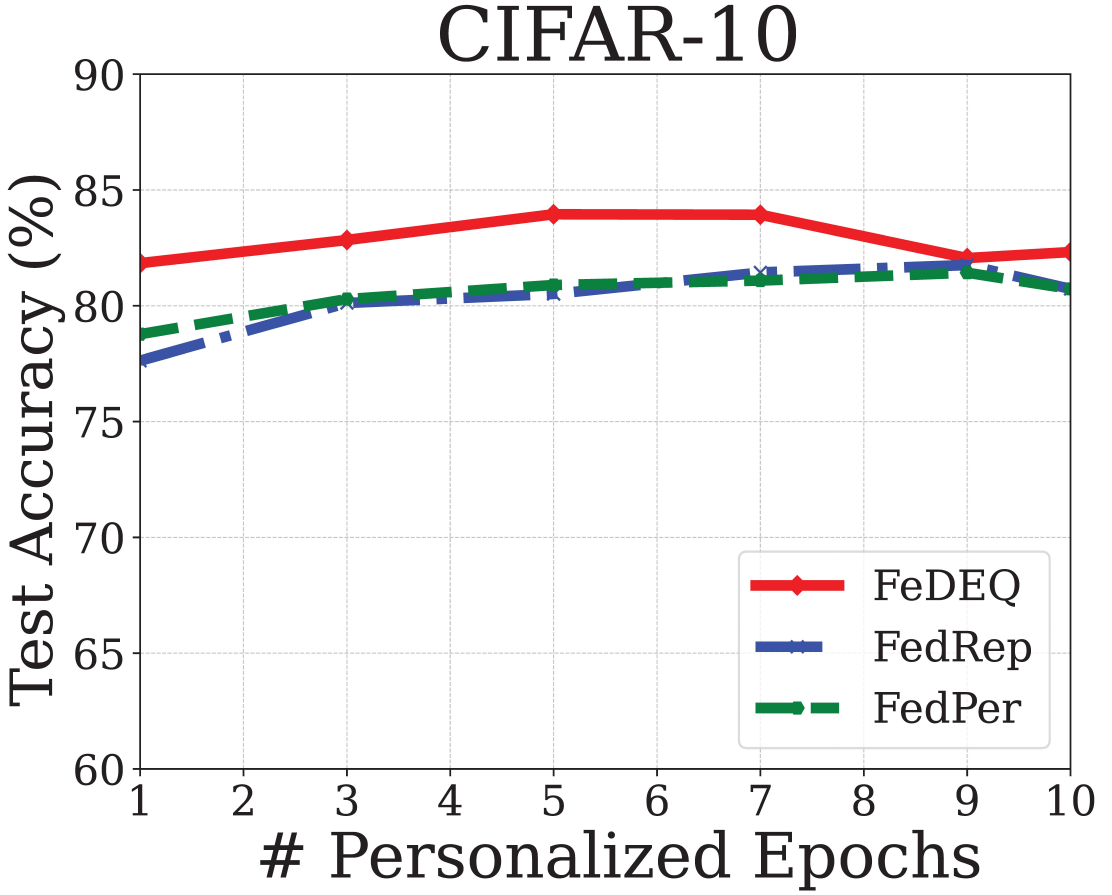}
        \vspace{-15pt}
		\caption{\label{fig:rep2}}
	\end{subfigure}
	\begin{subfigure}{.47\linewidth}
		\centering
		\includegraphics[width=\linewidth]{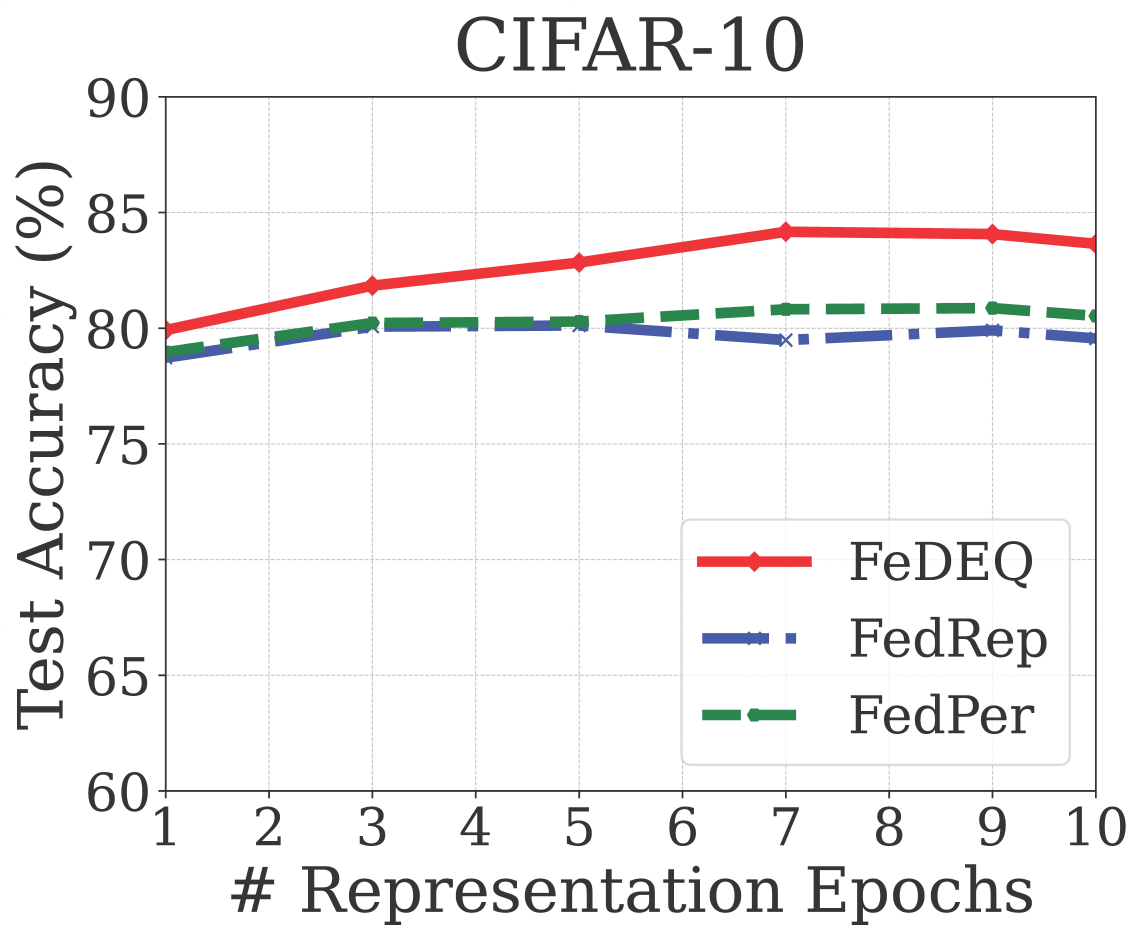}
        \vspace{-15pt}
		\caption{\label{fig:per2}}
	\end{subfigure}
    \vspace{-5pt}
	\caption{Effects of local epochs on learning global representation and personalized layers.}
	\vspace{-10pt}
\end{figure}

\textbf{Memory and Time Complexity:} To demonstrate \OurAlg's memory efficiency, we measure the average GPU memory consumption during training FedAvg with DEQ-ResNets on clients using CIFAR-100, in comparison to FedAvg with explicit ResNet models. Fig.~\ref{fig:memory} reveals that DEQ-ResNet-S, due to its smaller size, has the lowest memory footprint, saving twice as much memory as ResNet-14. Meanwhile, DEQ-ResNet-M's memory usage is moderately lower than that of ResNet-20 and is approximately 1.5 times less than ResNet-34 when training each batch of data. This can be attributed to DEQs' reliance on implicit differentiation, which eliminates the need to store intermediate values of all layers for backpropagation. However, the training time of \OurAlg is quite slower than that of explicit models due to the iterative nature of solving DEQ's forward pass. To mitigate this issue, one can explore acceleration methods like low-rank projection training~\citep{zhao2024galore, welore}.

\subsection{Ablation Study}
\noindent
\par
\textbf{Effects of $\rho$ on the Performance of \OurAlg:} 
The penalty parameter $\rho$ is a tuning parameter that controls the trade-off between the dual constraint satisfaction (i.e., alignment between local and shared representations) and the objective function minimization.
Exploring $\rho$ values of $\{0.001, 0.01, 0.1, 1.0, 5.0, 10.0\}$ on the FEMNIST dataset, we observed that \OurAlg converges across this range, as shown in Fig.~\ref{fig:rho_femnist}. However, a too-small $\rho$ weakens constraint enforcement, allowing local objectives more leeway but possibly deviating from the optimal consensus or slowing convergence. Meanwhile, a high $\rho$ value enforces strong consensus rapidly, which might limit local optimizations, leading to potentially suboptimal solutions by forcing agreement at the quality's expense. Empirical tests suggest $\rho = 0.01$ strikes the optimal balance for effectiveness and efficiency. It is worth noting that \OurAlg maintains convergence with similar performance, even with non-smooth activations like ReLU.

\textbf{Effects of Global Representation on Personalization:} We assess the representation and personalizer layers learned by \OurAlg, FedPer, and FedRep, particularly on  CIFAR-10, on two tailored scenarios. First, we vary the local epochs for updating personalized parameters while keeping a constant number of epochs for training the representation. As depicted in Fig.~\ref{fig:rep2}, \OurAlg achieves desirable results with fewer personalized updates than the others. Second, the emphasis is shifted towards the epochs for training the representation while the epochs for personalized parameter updates are kept constant. We observe that a satisfactory global representation could be achieved within a few epochs across all methods on both datasets, but \OurAlg potentially needs fewer to attain a good one, as shown in Fig.~\ref{fig:per2}. These results imply that \OurAlg can learn representations more effectively than its explicit counterparts.

%% file: CIKMSections/conclusion.tex
\section{Conclusion}

In this paper, we introduces \OurAlg, a unified FL framework addressing the challenges of data heterogeneity, communication, and memory constraints. 
Leveraging the compactness of deep equilibrium models and local fine-tuning for personalization, we formulate \OurAlg as an ADMM consensus optimization scheme that allows clients to learn a global equilibrium representation and then adapt to their personalized parameters. Extensive experiments on various benchmarks showcase \OurAlg achieves better generalization than SOTA methods with explicit models of up to 4$\times$ larger in size, highlighting \OurAlg's lower memory footprint during training, a direct benefit to FL for resource-constrained environments.
